\documentclass[10pt,twocolumn,letterpaper]{article}

\usepackage[pagenumbers]{wacv} % To force page numbers, e.g. for an arXiv version

\usepackage{graphicx}
\usepackage{amsmath}
\usepackage{amssymb}
\usepackage{booktabs}

\usepackage{graphicx}
\usepackage{amsmath}
\usepackage{amssymb}
\usepackage{booktabs}
\usepackage{enumitem}
\usepackage{tabularx}
\usepackage{graphicx}
\usepackage{amsmath}
\usepackage{amssymb}
\usepackage{booktabs}

\usepackage{graphicx}
\usepackage{amsmath}
\usepackage{amssymb}
\usepackage{booktabs}
\usepackage{times}
\usepackage{epsfig}
\usepackage{graphicx}
\usepackage{amsmath}
\usepackage{amssymb}
\usepackage{times,enumitem}
\usepackage{epsfig}
\usepackage{graphicx}
\usepackage{amsmath}
\usepackage{amssymb}
\usepackage{mathtools}

\usepackage[pagebackref,breaklinks,colorlinks]{hyperref}

\usepackage[capitalize]{cleveref}
\crefname{section}{Sec.}{Secs.}
\Crefname{section}{Section}{Sections}
\Crefname{table}{Table}{Tables}
\crefname{table}{Tab.}{Tabs.}

\def\wacvPaperID{1103} % *** Enter the WACV Paper ID here
\def\confName{WACV}
\def\confYear{2025}

\begin{document}

\title{Frame by Familiar Frame: Understanding Replication in Video Diffusion Models}

\author{Aimon Rahman $^*$, Malsha V. Perera $^*$, and Vishal M. Patel \\
Johns Hopkins University\\
$^*$ denotes equal contribution\\
{\tt\small \{arahma30,jperera4,vpatel36\}@jhu.edu}
% \and
% Malsha V. Perera \footnotemark[1]\\
% Johns Hopkins University\\
% {\tt\small jperera4@jhu.edu}
% \and
% Vishal M. Patel\\
% Johns Hopkins University\\
% {\tt\small vpatel36@jhu.edu}
}
% \maketitle

\twocolumn[{
\maketitle
\begin{center}
    \captionsetup{type=figure}
    \includegraphics[width=0.9\textwidth]{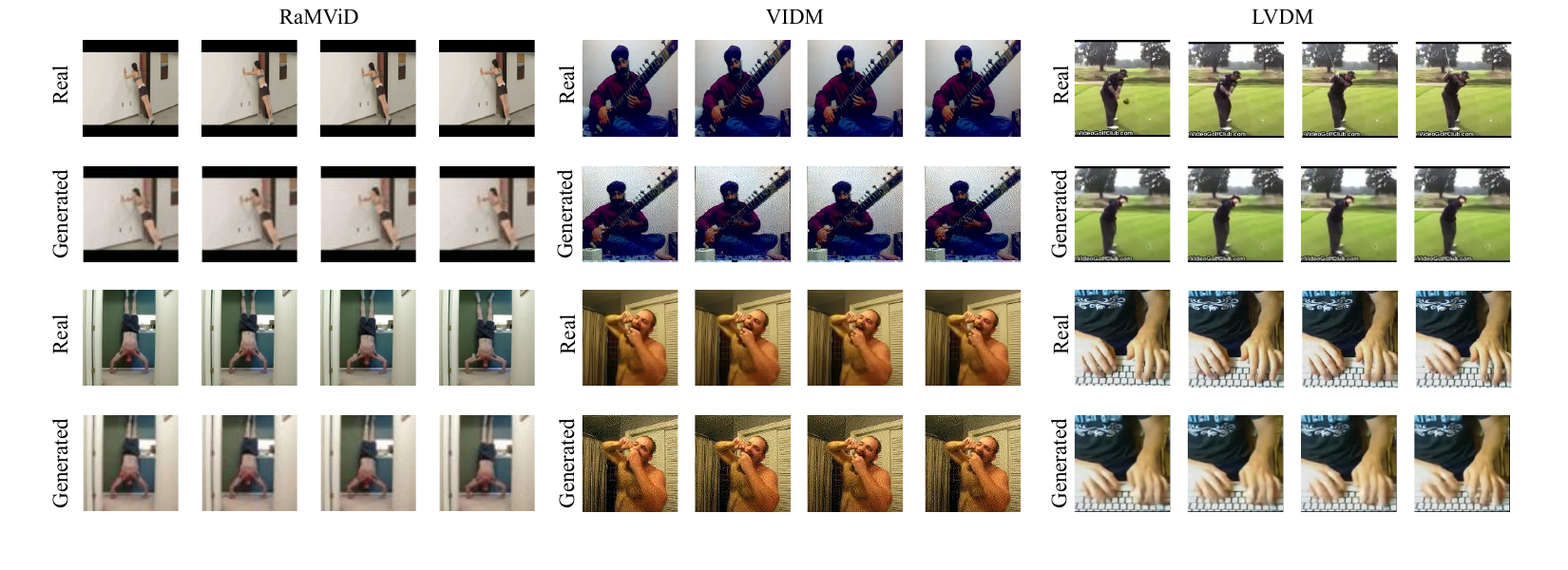}
   \vskip-15pt \captionof{figure}{Diffusion-based video synthesis models can sometimes replicate training data by assembling memorized foreground and background elements. We demonstrate this trend across multiple diffusion models trained on diverse datasets. Such occurrences prompt inquiries regarding data memorization and the ownership of videos produced by diffusion methods. Bottom row: Videos sourced from the RaMViD \cite{hoppe2022diffusion}, VIDM \cite{mei2023vidm}, and LVDM \cite{he2022latent} project websites. Top row: The most similar counterparts from the training dataset.}
    \label{intro}
\end{center}
}]

%%%%%%%%% ABSTRACT
\begin{abstract}
\vspace{-4mm}
   Building on the momentum of image generation diffusion models, there is an increasing interest in video-based diffusion models. However, video generation poses greater challenges due to its higher-dimensional nature, the scarcity of training data, and the complex spatiotemporal relationships involved. Image generation models, due to their extensive data requirements, have already strained computational resources to their limits. There have been instances of these models reproducing elements from the training samples, leading to concerns and even legal disputes over sample replication. Video diffusion models, which operate with even more constrained datasets and are tasked with generating both spatial and temporal content, may be more prone to replicating samples from their training sets. Compounding the issue, these models are often evaluated using metrics that inadvertently reward replication. In our paper, we present a systematic investigation into the phenomenon of sample replication in video diffusion models. We scrutinize various recent diffusion models for video synthesis, assessing their tendency to replicate spatial and temporal content in both unconditional and conditional generation scenarios. Our study identifies strategies that are less likely to lead to replication. Furthermore, we propose new evaluation strategies that take replication into account, offering a more accurate measure of a model's ability to generate the original content.
\end{abstract}

%%%%%%%%% BODY TEXT
\vspace{-3mm}
\section{Introduction}
\label{sec:intro}

Recent advancements in diffusion-based image generative models have paved the way for video generation \cite{ho2020denoising,song2020denoising,rombach2022high}. However, video synthesis has not reached the same prominence as its image counterpart, primarily due to the immense computational demands and the scarcity of expansive, public video datasets \cite{villegas2022phenaki}. Most current models limit themselves to producing short, low-resolution videos \cite{mei2023vidm,ho2022video}. The power of the video diffusion framework lies in its ability to utilize simple denoising networks that incorporate a temporal dimension for motion \cite{ho2022video,blattmann2023align}. Such models, when trained on video data, can generate realistic videos full of action and content. Yet, a pressing concern is the potential for training data replication. Even in the image domain, diffusion models are known to replicate content from training datasets, leading to concerns about the originality of the produced content \cite{somepalli2023diffusion,somepalli2023understanding}.

\noindent \textbf{Exploring Replication in Video vs. Image Generation Models.} The issue of training data replication is a well-documented challenge in image generation models, with significant research devoted to understanding its impacts and mechanisms \cite{somepalli2023diffusion,somepalli2023understanding}. These models demonstrate a remarkable ability to generate both unique content and, in some instances, partial or complete replications of their training data. In the domain of video generation models, this challenge is amplified due to the additional complexity of generating content that encompasses both static images and their temporal evolution with much smaller training data. Thus, it becomes imperative to examine the extent to which video generation models can innovate in terms of content and motion creation. Moreover, the field of video generation is diverse, covering areas such as video prediction, conditional and unconditional generation, text-conditioned generation, video infilling, etc. This underscores the necessity of investigating how video diffusion models manage the balance between replication and the generation of novel content.

\noindent \textbf{Implication of Data Replication in Video Diffusion Models.} The extent of data replication within video generation frameworks holds significant implications, beyond copyright violation, and in particularly in the domain of security and biometrics. A notable concern arises when a video replicates an individual's face from a training dataset, potentially leading to privacy issues. Moreover, a person's unique motion, such as their gait, can be distinct enough to facilitate identification \cite{klempousbiometric, kale2004identification,nixon2010human}. Replicating such motions, either by mimicking a person's gait or other physical patterns, might have detrimental effects on user authentication processes in behavioral biometrics. Additionally, the recent discovery of motion data for identifying individuals in virtual reality (VR) contexts amplifies these concerns \cite{nair2023unique,nair2023truth}. The implications of video replication thus extend beyond mere copyright infringement, especially when synthetic videos find applications in other downstream areas.\\ 
\noindent \textbf{Contributions.} Our research centers on the scientific inquiry into video diffusion models, examining their capacity for generating unique content, the extent and frequency of replication, and strategies for mitigating such replication. Our research delves into:
\begin{itemize}[label=-]
    \item Defining ``replication" in videos. This can be subjective, varying based on content diversity or viewer interpretation. We've pinpointed both clear and ambiguous instances of replication, differentiating between content and motion.
    \item Investigating the frequency of replication in video diffusion models, looking at both content and motion. We aim to determine if these models truly comprehend the actions they generate.
    \item Analyzing the relationship between the realism of generated videos and content replication. The hypothesis is that hyper-realistic videos might just be reflections of the training dataset.
    \item Examining video similarity metrics to effectively detect data replication. This will also help set benchmarks for future video diffusion model evaluations, especially since the current metrics, like the FVD, reward similarities to the training dataset.
    \item Offering recommendations for protocols in training and evaluating future video generation models. Our focus is on establishing guidelines that enhance model performance while ensuring diverse and original content generation, moving beyond current practices that might inadvertently favor replication over innovation.

\end{itemize}

\section{Related Work}

Our study intersects with various domains: diffusion models, image/video generation techniques, and the inclination of generative models to mimic samples. Here, we briefly outline the foundational concepts from each area, emphasizing their interrelations and addressing the challenges of video sample replication.

\noindent \textbf{Sample Replication in Generative Models.} Recent research has highlighted a trend of training data replication or memorization in popular generative models such as Generative Adversarial Networks (GANs) and diffusion models. As these models become increasingly adept at creating hyper-realistic images, questions arise about the originality of these images versus their being mere duplications from training datasets. The ``this person does not exist" phenomenon, famously associated with faces generated by StyleGAN \cite{karras2019style}, has been scrutinized, revealing the potential to trace back to the original dataset used in the model \cite{webster2021person}. It is observed that the tendency of GANs to replicate training data decreases exponentially with the increase in dataset complexity and size \cite{feng2021gans}. The suggestion has been made that the duplication of training data is a significant factor in this context. By employing non-parametric tests, it is possible to address and mitigate this issue effectively \cite{meehan2020non}. This phenomenon of replication extends to diffusion models too, where retrieving training data from the model is possible \cite{carlini2023extracting}. Although often attributed to training on smaller datasets, this replication behavior is also evident in models trained on larger datasets \cite{somepalli2023diffusion,somepalli2023understanding}. The issue becomes particularly concerning in terms of social bias, especially with diffusion models trained on facial datasets, where replication behavior is prominent \cite{perera2023analyzing}.

\noindent \textbf{Diffusion-based Video Generation.} Diffusion Probabilistic Models (DPMs) are a subset of deep generative models known for their ability to incrementally introduce noise into data points \cite{ho2020denoising,song2020denoising}. This process is followed by a denoising phase that iteratively cleans the data, resulting in the generation of new samples. DPMs have shown remarkable results in producing high-quality and diverse images, inspiring researchers to explore their potential in video generation, prediction, and interpolation \cite{ho2022video,luo2023videofusion,voleti2022mcvd,hoppe2022diffusion,ho2022imagen}.

Applying DPMs to video generation is still in its infancy and presents unique challenges \cite{villegas2022phenaki}. Videos possess higher-dimensional data and intricate spatiotemporal relationships, making the task more complex. Diffusion-based video generation models operate on the principles similar to image diffusion models, with a key distinction in their architecture\cite{ho2022video}. These models typically use either a 3D Unet architecture \cite{ho2022video,hoppe2022diffusion}, which adds depth to the processing, or a conditional 2D Unet that takes into account the previous frame in a sequence \cite{voleti2022mcvd}. Most of these models incorporate a temporal layer to capture motion, and variations of this layer are evident in the current research \cite{molad2023dreamix,blattmann2023align,luo2023videofusion,mei2023vidm}. The process of generating new frames in these models is autoregressive; the creation of each subsequent frame depends on the preceding one. This can be either conditional or unconditional. In a conditional generation, some models use textual descriptions to guide the generation process \cite{yin2023nuwa,villegas2022phenaki,molad2023dreamix}. Others might use the first frame or a few initial frames as a basis for what is essentially video prediction \cite{hoppe2022diffusion,voleti2022mcvd}. When the generation is steered by intermittent frames throughout the video, it is referred to as video infilling. Unconditional generation, on the other hand, relies purely on noise to create videos.
Regardless of the approach, these models require an intrinsic understanding of motion and content to perform their tasks effectively. It is also worth noting that these video generation models are often \textit{not publicly available}. They can be resource-intensive, both in terms of training and testing, which might contribute to their limited accessibility.\\
\noindent \textbf{Video copy detection and localization.}
Video Copy Detection (VCD) involves identifying pairs of query-reference videos containing copied content without localizing the common content within the videos. In contrast, Video Copy Localization (VCL) requires finding the exact temporal segments within a pair of videos that contain duplicated content. VCD can be performed either at the video-level or frame-level. Video-level approaches utilize standard similarity measures on global representations of the videos to identify copied content \cite{Zilos2017near,Cai2011mil, Song2011multi}. However, these approaches are less effective in partial copy detection tasks as they aggregate with irrelevant content and clutter. Meanwhile, frame-level features with spatio-temporal representations have proven to yield a significant advantage in video retrieval tasks and are advantageous in precisely locating copied segments. Various techniques, such as Fourier-based representations \cite{baraldi2018lamv}, multi-attention networks \cite{wang2021attent}, or transformer-based networks \cite{He2022learn,Shao_2021_WACV}, are used for temporal aggregation. Recent works employ a video similarity network that captures fine-grained spatial and temporal structures within pairwise video similarity matrices \cite{zilos2019visil}. In VCL, frame-level feature representations are followed by a temporal alignment module that needs to reveal the time range of one or multiple copied segments between the potential copied video pair. Frame-level features can typically be extracted using image descriptors commonly employed in image copy detection. One popular example is SSCD \cite{pizzi2022self}, an image descriptor based on self-supervised learning, which optimizes a descriptor for copy detection through entropic regularization. Temporal Hough voting \cite{douze2010compact}, temporal networks \cite{tan2009scal}, and dynamic programming \cite{chou2015pattern} are examples of some of the simplest VCL methods.

\begin{figure}[t!]
\centering
\includegraphics[width=0.7\linewidth,scale=1]
{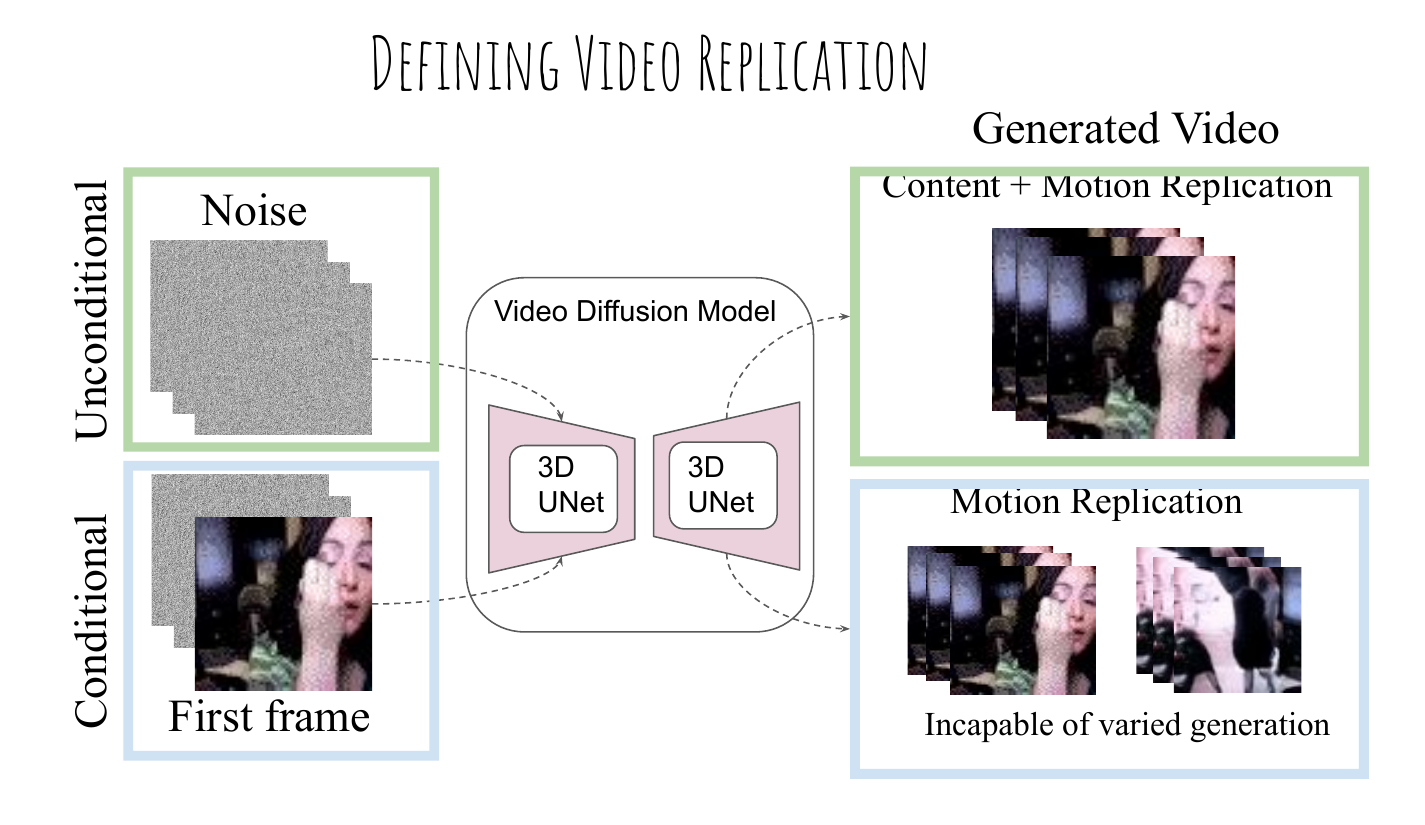}
\vskip-10pt
\caption{Definition of replication in video generation domain. Content and motion replication refers to the direct duplication of content and motion from the training dataset, essentially producing a 1:1 copy. On the other hand, motion replication assesses a video generation model's inherent ability to create motion from an initial frame. This initial frame supplies the content context, but the true measure of a video generation network's capability lies in its understanding of the comprehensive content within that first frame. The critical question is whether the network genuinely comprehends and generates subsequent motion, or if it merely replicates sequences it has learned from the training data.}
\label{Fig:intro-fig}
\vspace{-4mm}
\end{figure}

\section{Defining Video Replication}
Replicating content in the imaging domain generally means that a significant portion of the training image appears in the generated images \cite{somepalli2023diffusion}. What's considered a ``significant portion" varies, but it often refers to an easily recognizable region. In essence, it is considered a generated image to have duplicated content if it features an object (either in the foreground or in the background) that mirrors an object from a training image, allowing for slight variations that might arise from data augmentations \cite{somepalli2023diffusion}.

However, defining data replication in video generation is a tad more intricate. For unconditional generation, which is the generation from pure random noise without any specific guiding condition, data replication can encompass both the \textit{subject and its motion}. In the domain of conditional generation, where the model is given an initial frame and then predicts the subsequent video, the question arises: \textit{Does the model genuinely understand and generate motion, or is it merely recalling patterns from the training dataset?} This necessitates different definitions of video replication based on the generation context (conditional or unconditional) as illustrated in Figure \ref{Fig:intro-fig}.

For unconditional generation, if the content closely matches that of a training video with minimal differences, it is regarded as replication. Essentially, the replication of visual elements (content) from the training video inevitably leads to the replication of the dynamics or movement patterns (motion) present in the original video. On the other hand, in the conditional context, if a model reproduces the exact movement sequence after being provided with just an initial frame, it is seen as motion replication. In the latter scenario, it is understandable if a model replicates motion when predicting or infilling video. However, to test genuine understanding, alterations can be made to the initial frames—like changing the angle or occluding the frame—to see if the model can still predict plausible subsequent motion. If the model simply replicates the training data to generate motion based on the first frame, this process is referred to as \textit{motion replication} in this paper.

\vspace{-3mm}
\section{Detecting Data Replication in Video Diffusion Models}
\subsection{Content Replication.} 
\label{content replication}
Content replication in the context of text-to-video or unconditional video generation refers to a scenario where the frames generated by the model contain the same or strikingly similar content to what it has seen during training. This means that instead of creating new, original content based on the learned concepts and dynamics, the model reproduces specific examples from its training data, which suggests a lack of true generative capacity or understanding. Due to limited access to publicly available models, our analysis is mostly based on the generated samples showcased on the official websites of the respective papers. We specifically focus on results that are obtained through unconditional generation, ensuring that initial frame contents don't influence the generated samples.\\
\noindent \textbf{VSSCD Score.} Initial findings indicate that a significant portion of the video samples generated unconditionally by the model are direct replicas of the training videos, featuring identical content. To identify these replications, we adapted the Self-Supervised Copy Detector (SSCD) \cite{pizzi2022self}, originally designed for image copy detection, for use with video content. Our methodology involves segmenting frames from reference videos, extracting their features, and then concatenating these features. This process is repeated for all training videos. Next, we employ SSCD to extract features from these concatenated frames, denoting these video features as VSSCD. We then calculate the cosine similarity to identify the top matches. Let's denote \( R = \{R_1, R_2, ..., R_n\} \) as the set of real videos and \( G = \{G_1, G_2, ..., G_m\} \) as the set of generated videos. We calculate the similarity between each pair of VSSCD features of real and generated videos. The similarity between a real video \( R_i \) and a generated video \( G_j \) is denoted as \( \text{VSSCD}(R_i, G_j) \). The process can be represented by the following equation:
% \[
% \text{Top-VSSCD} = \max \left\{ \text{VSSCD}(R_i, G_j): 1 \leq i \leq n, \, 1 \leq j \leq m \right\}
% \]
\setlength{\belowdisplayskip}{0pt} \setlength{\belowdisplayshortskip}{0pt}
\setlength{\abovedisplayskip}{0pt} \setlength{\abovedisplayshortskip}{0pt}
\begin{multline}
    \text{Top-VSSCD} = \max \bigl\{ \text{VSSCD}(R_i, G_j): \\
    1 \leq i \leq n, \, 1 \leq j \leq m \bigr\},
\end{multline}
where \( \text{Top-VSSCD} \) is the highest similarity score among all combinations of real and generated videos. In this work, we use the term `VSSCD score' to refer to the similarity score between the VSSCD features of two videos. The validity of this VSSCD-based approach is demonstrated in Table \ref{valid}, where we present the VSSCD scores for exact copies, various augmented copies, and scores corresponding to random videos that do not match the reference.

\begin{table}[h!]
  \centering
  \caption{VSSCD ($\downarrow$) Scores for Replication Detection in the UCF-101 \cite{soomro2012ucf101} and Kinetics-400 \cite{kay2017kinetics} Datasets. The table compares the VSSCD scores for exact 1:1 copies, augmented copies, and unrelated videos, demonstrating the reference efficacy of VSSCD in identifying replicated content in videos.}
  \resizebox{\columnwidth}{!}{
    \begin{tabular}{
        l|
        c|
        c|
        c |
        c |
        c |
        c |
        c }
      Frame Operation & 1:1 & Flip & Crop & Occlusion & Translation & Rotation & Random \\ \hline
      UCF-101 \cite{soomro2012ucf101}  & 1 & 0.9684 & 0.9032 & 0.9998 & 0.9147 & 0.8574 & 0.0788 \\ 
      Kinetics-400 \cite{kay2017kinetics} & 1 & 0.9800 & 0.9026 & 0.9999  & 0.9043 & 0.8031 & 0.1046 \\
      \hline
    \end{tabular}
  }
  \label{valid}
\end{table}

 We evaluate the extent of sample replication in various video generation models using VSSCD score and compare these findings with their Fréchet Video Distance (FVD) scores. We examine the output of several models including VIDM \cite{mei2023vidm}, VDM \cite{ho2022video}, RaMViD \cite{hoppe2022diffusion}, and LVDM \cite{he2022latent}. For this analysis, we only train the VDM model directly on the UCF-101 dataset to generate videos. For the other models - VIDM, RaMViD, and LVDM - we use the video outputs available on their respective websites. All of these models were initially trained on the UCF-101 \cite{soomro2012ucf101} dataset. Note that, the generation is unconditional, hence content and motion are generated from pure noise.

 Furthermore, as shown in Table \ref{quanres}, a high VSCCD score, indicating greater similarity to the training set, corresponds to an improved FVD score. This occurs when the generated videos contain more replicas of the training data, as the FVD score tends to assign lower values to these replicas, rewarding exact replication rather than novel content generation. This is a limitation of relying solely on the FVD score. To address this issue, we propose complementing the FVD results with VSSCD scores. The details of this integrated approach are discussed in Section \ref{FVD-VSSCD Curve}.

\begin{table}[h!]
  \centering
  \caption{Quantitative comparison of the FVD ($\downarrow$) and VSSCD ($\downarrow$) scores among various video generation networks.}
  \resizebox{\columnwidth}{!}{
    \begin{tabular}{
        l| 
        c |
        c |
        c |
        c }
      Metrics & VDM \cite{ho2022video} & VIDM \cite{mei2023vidm} & LVDM \cite{he2022latent} & RaMViD \cite{hoppe2022diffusion}   \\ \hline
     VSSCD & 0.591 & 0.6347 & 0.694 & 0.744 \\ 
      FVD \cite{voleti2022mcvd} & 631 & 172.77 & 151.34 & 152.24 \\
      \hline
    \end{tabular}
  }
  \label{quanres}
  \vspace{-6mm}
\end{table}

\begin{figure*}[h!]
\centering
\includegraphics[width=0.7\linewidth,scale=1]
{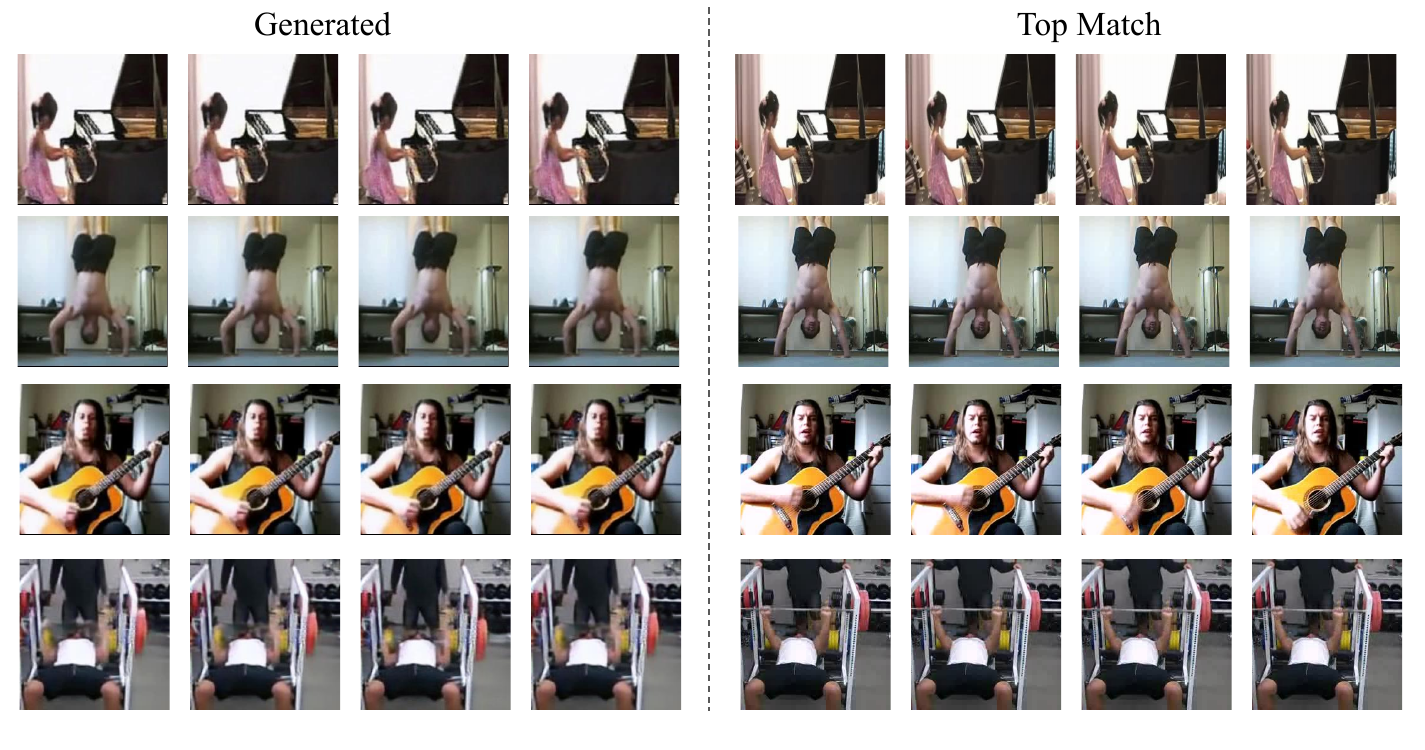}

\vskip-14pt\caption{The highest similarities identified within the UCF-101 dataset compared against outputs generated by an unconditional video generation approach, utilizing a latent video diffusion model (LVDM) \cite{he2022latent}.}
\label{Fig:sim}
\vspace{-4mm}
\end{figure*}

\noindent \textbf{Discussion.}  We found a tendency among these models to produce videos that replicate training data. This is reflected in Table \ref{quanres}. A shared trait among these models is their training on smaller or limited datasets, such as UCF-101 \cite{soomro2012ucf101} for VIDM \cite{mei2023vidm}, VideoFusion \cite{luo2023videofusion}, RaMViD \cite{hoppe2022diffusion}, etc. This leads us to assume that models trained on limited datasets might be more prone to produce replicated videos due to their limited content understanding. This is evident for both latent and regular video diffusion models, as seen in Figure \ref{intro} and Figure \ref{Fig:sim}. This phenomenon is not exclusive to the video domain; similar trends have been observed in image generation models \cite{somepalli2023diffusion,somepalli2023understanding}. However, the extent of these trends in image models is not as pronounced as in the video domain. Moreover, what constitutes a ``small dataset" varies between the two domains, as video data has to provide both content and motion dynamics. 

% In the video generation domain, a dataset not only conveys content but also imparts motion dynamics, offering a richer, multidimensional source of information for generative models.

\vspace{.5cm}
\noindent\fbox{%
    \parbox{0.47\textwidth}{% Adjust the width as needed
        \textbf{\textcolor{blue}{Observation.}} Video diffusion models trained from scratch on limited video datasets
exhibit a greater tendency to completely replicate the content of the videos within the
training dataset.

    }%
}

\begin{figure*}[h!]
\centering
\includegraphics[width=1\linewidth,scale=1]
{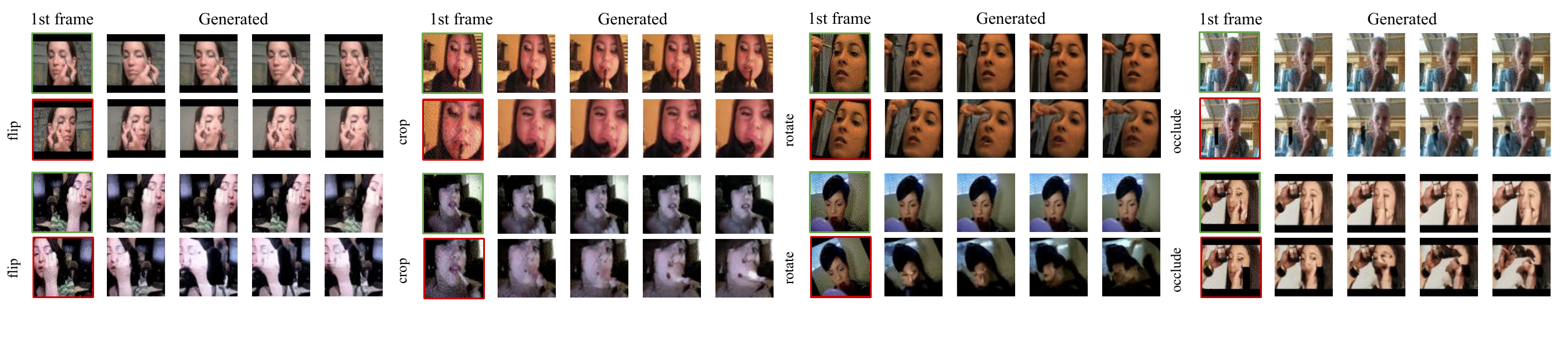}

\vskip-15pt\caption{The initial frame provided as a condition to the video generation model is denoted as the `$1^{st}$ frame'. `1st Frames' represent their original orientation from the dataset, while those with a red outline signify altered frames. Observations show that the model properly generates motion when presented with frames in their original orientation. However, it struggles to produce consistent motion when given an augmented version of the same image, indicating the model memorized the motion.}
\label{Fig:rep}
\vspace{-7mm}
\end{figure*}

\subsection{Motion Replication} 
In video prediction tasks, distinguishing between genuine generation and mere replication is challenging, especially when the model is given initial frames to predict subsequent motion. While it is straightforward to spot replication in unconditional generation—by identifying content that mirrors the training set—conditional generation complicates detection. Theoretically, a model with a deep understanding of motion should produce diverse outcomes from the same frame, given the many motion trajectories possible. Producing a singular, training set-specific trajectory might indicate replication, but arguing against it is tough since it might be deemed the `best' result. To probe the model's grasp of motion, we utilized pre-trained video diffusion models \cite{hoppe2022diffusion,voleti2022mcvd} on various datasets.

To evaluate whether a video generation or prediction model possesses a genuine comprehension of the motion dynamics or is merely replicating the learned patterns, we subject the model to a test using the original dataset's initial frame along with its variants—flipped, cropped, occluded, rotated, and translated. We then assess the model's performance by comparing the Fréchet Video Distance (FVD) \cite{unterthiner2019fvd} of the output based on the original first frame to that of the augmented versions. We hypothesize that, ideally if the model truly understands motion, the FVD scores should be relatively consistent across all variations.

\begin{table}[h!]
	\centering
		\caption{Quantitative comparison of FVD ($\downarrow$) scores on UCF-101 \cite{soomro2012ucf101} and Kinetics-400 \cite{kay2017kinetics} datasets among various video prediction networks\cite{hoppe2022diffusion}.
  }

	\resizebox{0.5\columnwidth}{!}{
		\begin{tabular}{@{\extracolsep{\fill}}
				l 
				c 
				c }
			{Frame Orientation} & \multicolumn{2}{c}{Video Prediction Model \cite{hoppe2022diffusion}} \\ \hline
			& UCF-101 \cite{soomro2012ucf101}  & Kinetics-400 \cite{kay2017kinetics} \\ \hline

			Original & 667.64 & 824.99\\ 
   \hline
     	Flip & 942.02 & 916.22 \\ 
					Crop  & 896.46 & 981.83 \\ 
   
			Occlusion  & 867.57 & 956.31 \\

			Translation  & 873.36 & 900.33 \\ 
            
			Rotation  & 806.02 & 1022.34\\ 
			\hline
			
		\end{tabular}
	}

	\label{flip-aug}
 \vskip -15pts
\end{table}

\noindent \textbf{Discussion.}   Our results indicate that the model, when provided with an initial frame in its original orientation, is capable of generating subsequent frames effectively. However, it struggles to maintain this performance when presented with minor alterations, such as flipping or cropping, which do not alter the frame's semantic content. This suggests a tendency towards overfitting on the training dataset. As demonstrated in Table \ref{flip-aug}, the original orientation consistently results in superior FVD scores, with any form of variation leading to a degradation in performance. This observation is further supported qualitatively, as illustrated in Figure \ref{Fig:rep}.

\vspace{2mm}
\noindent\fbox{%
    \parbox{0.47\textwidth}{% Adjust the width as needed
        \textbf{\textcolor{blue}{Observation.}} Video prediction models frequently memorize the motion dynamics present in the training dataset, resulting in a limited ability to generate novel motion patterns.

    }%
}

\section{Replication in Video Diffusion Models}

\subsection{Effect of dataset size on video replication}
Previous studies on diffusion-based image generation have demonstrated that data replication is more prevalent when models are trained on smaller datasets. To investigate whether diffusion-based video generation exhibits the same behavior, we conducted the following experiment. We trained multiple VDM-based video diffusion models \cite{ho2022video} on randomly sampled subsets of the UCF-101 dataset \cite{soomro2012ucf101}. The subsets consisted of 100, 300, 500, 700, and 1000 videos. As shown in Table \ref{tab:train size}, when the dataset size decreases, diffusion models tend to replicate more frequently, as indicated by the higher VSSCD scores, which reflect a greater degree of replication. Therefore, similar to image generation, video diffusion models also show a higher tendency to replicate data when trained on smaller datasets.

\begin{table}[h]
\centering
\caption{Results on training data size vs. replication as measured by VSSCD ($\downarrow$) score}
\resizebox{0.9\columnwidth}{!}{
\begin{tabular}{cccccc}
\hline
\textbf{Training data size (samples)} & 100 & 300 & 500 & 700 & 1000 \\ \hline
\textbf{VSSCD Score} & 0.9468 & 0.9142 & 0.8976 & 0.8639 & 0.8251 \\ \hline
\end{tabular}
}
\label{tab:train size}
\vskip -15pts
\end{table}

\subsection{Replication in Text-to-Video models}

\begin{figure}[t!]
\centering
\includegraphics[width=0.7\linewidth,scale=1]
{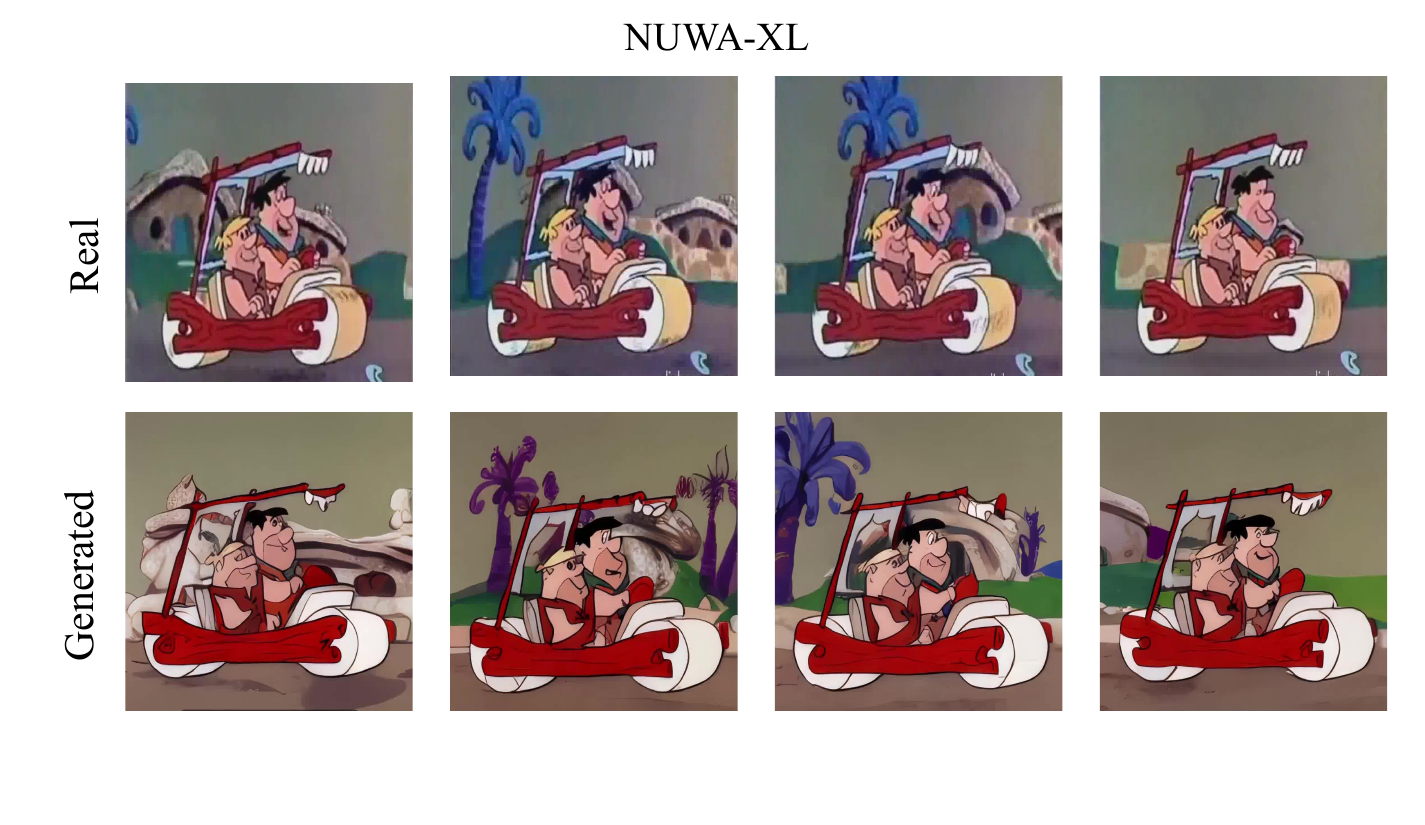}

\vskip-14pt\caption{An instance of replication in a text-to-video (T2V) model \cite{yin2023nuwa}. Generated with the text prompt ``Fred and Barney driving a car". NUWA-XL has been trained solely on the episodes of ``The Flintstones". The replicated segment is from the episode ``Disorder in the Court".}
\label{Fig:t2v-reps}
\vspace{-8mm}
\end{figure}

\textbf{Large Scale T2V diffusion model.} The foundation for many text-to-video (T2V) diffusion models \cite{blattmann2023align,wang2023modelscope,luo2023videofusion,zhou2022magicvideo,zhang2023i2vgen} has been the WebVid-10M dataset \cite{bain2021frozen}. The trend of sample replication in large-scale video diffusion models trained on extensive datasets, such as WebVid-10M, is notably less pronounced. Table \ref{tab:VSSCD_scores} presents the average top VSCCD scores generated by these models. These scores indicate a lower tendency for video data replication, particularly when compared to models trained from scratch on smaller datasets. The backbone model's initial capabilities in image generation, enhanced through exposure to vast video data, likely contribute to this reduced replication and boost creative outputs. It is important, however, to recognize that these models, often built upon the principles of Stable Diffusion \cite{Latentdiff} for text-to-image tasks, are not completely immune to replicating samples from their own image training datasets \cite{somepalli2023diffusion, somepalli2023understanding}.

\begin{table}[h]
\centering
\caption{VSSCD ($\downarrow$) Scores for various large-scale T2V diffusion models trained on WebVid-10M.}
\vspace{-3mm}
\resizebox{0.4\columnwidth}{!}{
\begin{tabular}{cc}
\hline
\textbf{Model} & \textbf{VSSCD Score} \\ \hline
ModelScope\cite{wang2023modelscope}     & 0.39                 \\ \hline
VideoCrafter \cite{chen2023videocrafter1}   & 0.22                 \\ \hline
ZeroScope \cite{khachatryan2023text2video}      & 0.35                 \\ \hline
AnimateDiff \cite{guo2023animatediff}   & 0.22                 \\ \hline
\end{tabular}
}
\label{tab:VSSCD_scores}
\vskip -15pts
\end{table}

Additionally, we conducted experiments to examine the impact of using a pre-trained network and its relationship to replication. Table \ref{tab:pretrain} presents various training settings applied to the pre-trained network from \cite{wang2023modelscope}. The results show that fine-tuning the pre-trained network significantly reduces the replication score.

\begin{table}[h]
\centering
\caption{Pre-training Results on Video Diffusion Models. Pre-trained models are trained on WebVid-10M and finetuned on UCF-101.}
\vspace{-5mm}
\resizebox{0.6\columnwidth}{!}{
\begin{tabular}{cc}
\hline
\textbf{Model} & \textbf{VSSCD Score ($\downarrow$)} \\ \hline
Scratch Trained     & 0.74                \\ \hline
Pre-trained + Fine-tuned Temporal only  & 0.30                 \\ \hline
Pre-trained + full Fine-tuned      & 0.47                 \\ \hline
Pre-trained + LoRA fine-tuned  & 0.37                 \\ \hline
\end{tabular}
}
\label{tab:pretrain}
\vskip -15pts
\end{table}

% Unfortunately, this dataset has been retracted due to copyright issues. The construction and operation of these expansive T2V models not only demand substantial computational resources but also face availability restrictions, with most not being accessible to the public. This presents significant challenges for those aiming to replicate these models, as access to both the datasets and the models themselves is restricted.

% Nonetheless, in Section \ref{t2i}, we demonstrate that extending models from the T2I architecture results in a reduced propensity for direct replication, owing to the image generation model's inherent capability for creative output.

\noindent \textbf{T2V Model trained from Scratch.}
 In this section, we evaluate the T2V model trained on a large dataset, with no T2I backbone. For our experiment, we selected the text-to-video model NUWA-XL \cite{yin2023nuwa}. While the model itself is not publicly available, videos generated by it are available. Notably, NUWA-XL has been trained on the episodes of ``The Flintstones". Given that ``The Flintstones" consists of 166 episodes, each approximately 25 minutes long, this constitutes a sizeable dataset for analysis. Our observations revealed instances of replication in NUWA-XL, as illustrated in Figure \ref{Fig:t2v-reps}. While these occurrences of replication are less frequent than (~47\% average top VSSCD) in unconditional models, they underscore that text-to-video models are not entirely immune to replication phenomena.

 \vspace{-2mm}
\subsection{Data Requirements for Unique Content: Image vs. Video Diffusion Models}

\begin{figure}[h!]
\vspace{-10pt}
\centering
\includegraphics[width=0.7\linewidth]
{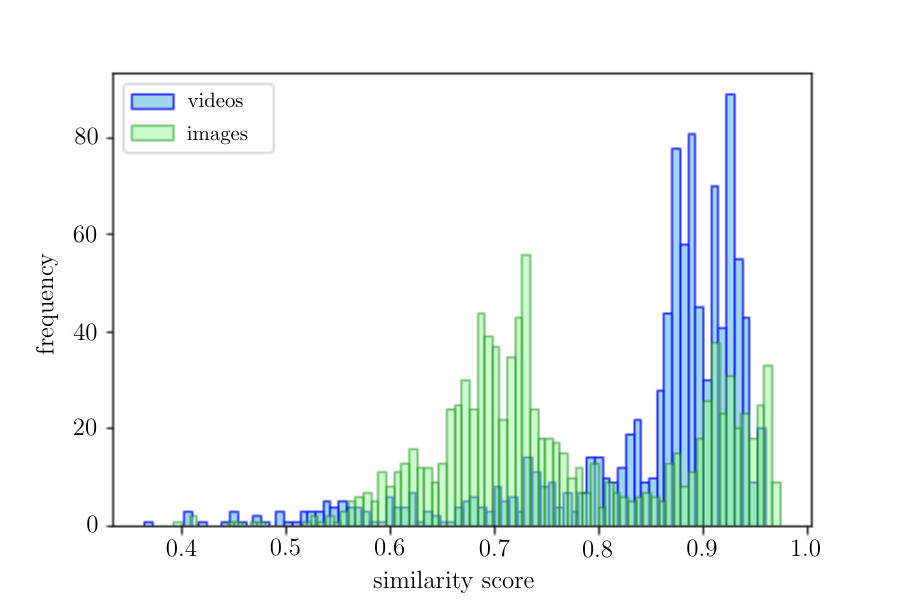}
\vskip-10pt\caption{The histogram illustrates the similarity scores between the generated images or videos and their corresponding training data. Specifically, the similarity score for the generated image samples from the image-based diffusion model is determined by the cosine similarity between the SSCD features of the generated image and the best-matched frame used during model training. Similarly, for samples generated from the video-based diffusion model, the calculation of similarity scores follows the procedure outlined in Section \ref{content replication}.}
\label{Fig:histogram}
\vspace{-10pt}
\end{figure}

In this section, we draw comparisons between sample replication issues in video diffusion models and their image generation counterparts. We initiate the process by training an image generation diffusion model \cite{ho2020denoising} using individual frames extracted from videos, identifying instances where the model generates unique images. Subsequently, we apply a similar training regimen to a video diffusion model \cite{ho2022video} and identify unique video outputs. Our primary focus is to investigate the data requirements for each model type to produce original results. Through our experiments, we find that the image generation model is capable of generating unique samples with as few as 1,000 data points. In contrast, this same amount of data proves to be insufficient for the video model to achieve a similar level of unique output generation. This is illustrated in Figure \ref{Fig:histogram}.

\noindent\fbox{%
    \parbox{0.47\textwidth}{% Adjust the width as needed
        \textbf{\textcolor{blue}{Observation.}} 
When training from scratch, the amount of data needed for generating unique videos is significantly greater than that required for the image generation models.

    }%
}

\section{Mitigating Video Replication: Recommended Protocls}

% Some approaches to video diffusion model training involve building the model from scratch using video data. Alternatively, there's a strategy that capitalizes on the strengths of pre-existing image generation models \cite{blattmann2023align,wang2023modelscope,villegas2022phenaki}. 

Video diffusion models mentioned in this paper can differ significantly in their training approaches, architectures, and dataset sizes. Crafting a customized solution for each model type may be unrealistic and exceeds the scope of this study. Instead, we present a set of guidelines aimed at evaluating video diffusion models, particularly emphasizing the assessment of their ability to replicate outcomes. Furthermore, we suggest strategies for training these models on small datasets, which is particularly relevant for minimizing issues of replication in scenarios where resources are limited.

\subsection{The Integrated FVD-VSSCD Curve.}
\label{FVD-VSSCD Curve}
The Fréchet Video Distance (FVD) \cite{unterthiner2019fvd} is a commonly employed metric for evaluating video generation models. FVD operates by extracting features from both generated and real videos using a pre-trained feature extractor, typically an I3D model \cite{carreira2017quo} trained on the Kinetics-400 dataset \cite{kay2017kinetics}. It then calculates the mean and covariance of these feature distributions for both real and generated videos and computes the distance between them \cite{eiter1994computing}. Ideally, a smaller distance signifies better performance. However, there is an inherent flaw in this metric -- if the generated videos are exact replicas of the training data, the distance diminishes, inadvertently rewarding exact replication rather than novel content generation. 

% We can also see this in Table \ref{quanres}, where a high VSSCD score, indicating greater similarity to the training set, corresponds to a better FVD.

\begin{figure}[h!]
\centering
\includegraphics[width=0.7\linewidth]
{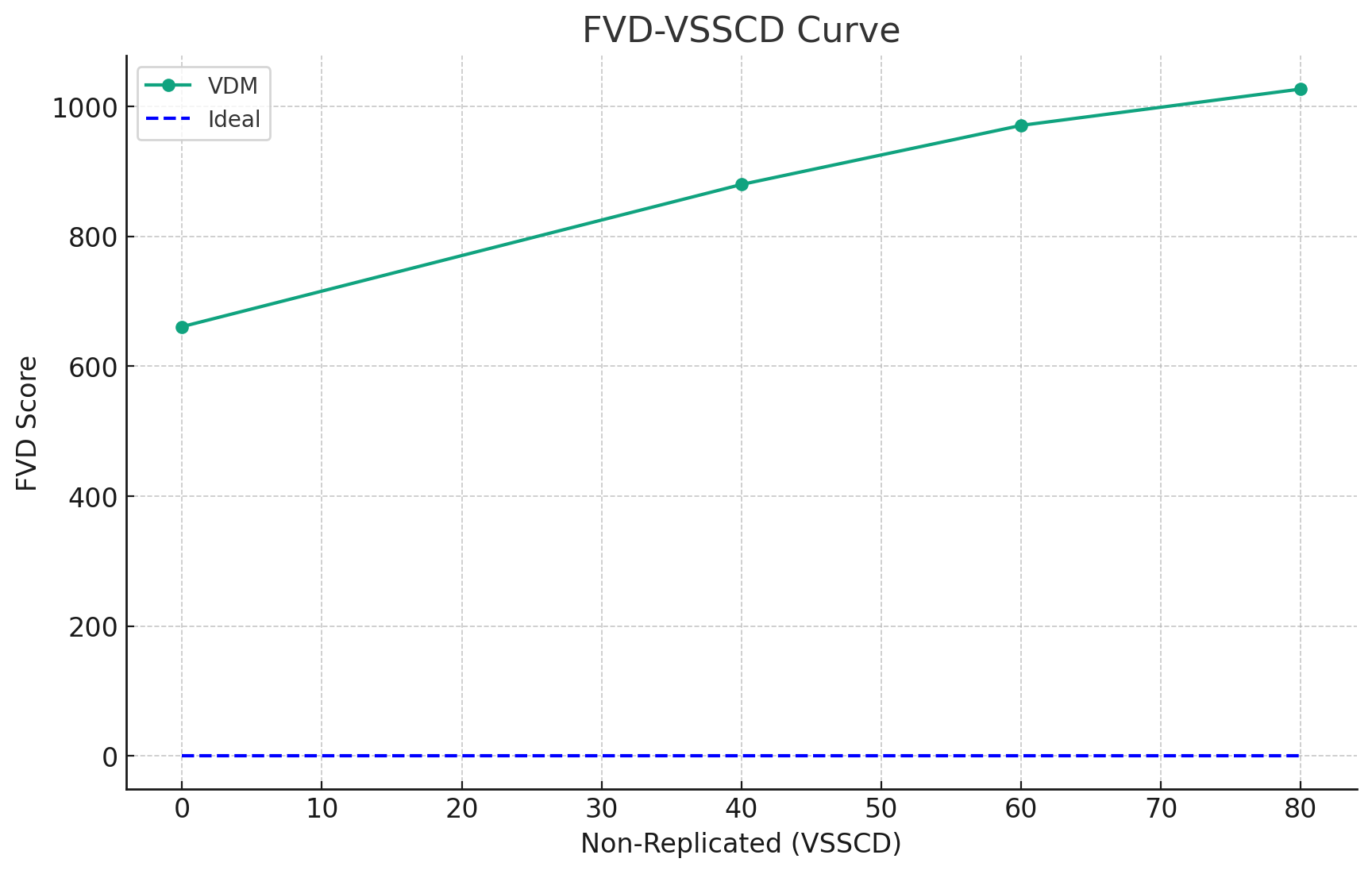}

\vskip-10pt\caption{A representation of the FVD-VSSCD Curve. The y-axis represents the FVD score, which measures the quality of video frames. The x-axis indicates the percentage of non-replicated samples, which reflects the proportion of the generated videos that have been filtered out based on a certain threshold of the VSSCD score (0.6). }
\label{Fig:SSD-FVD}
\vspace{-7pt}
\end{figure}

To address this limitation, we suggest complementing the FVD results with the VSSCD scores. Our approach involves not just reporting the FVD scores but also recalculating them after excluding the generated videos that are replicates of the training content. Specifically, we remove the generated videos exhibiting various degrees of similarity to the training data and then recalculate the FVD. In an ideal scenario, the resulting graph of the FVD scores versus the percentage of removed replicated samples would display a consistent trend. This would indicate that the videos are not only realistic but also distinct from the training data. We present this analysis in Figure \ref{Fig:SSD-FVD}, illustrating the relationship between the FVD and replication rates.
\vspace{-3mm}

\begin{figure}[h!]
\centering
\includegraphics[width=0.8\linewidth,scale=1]
{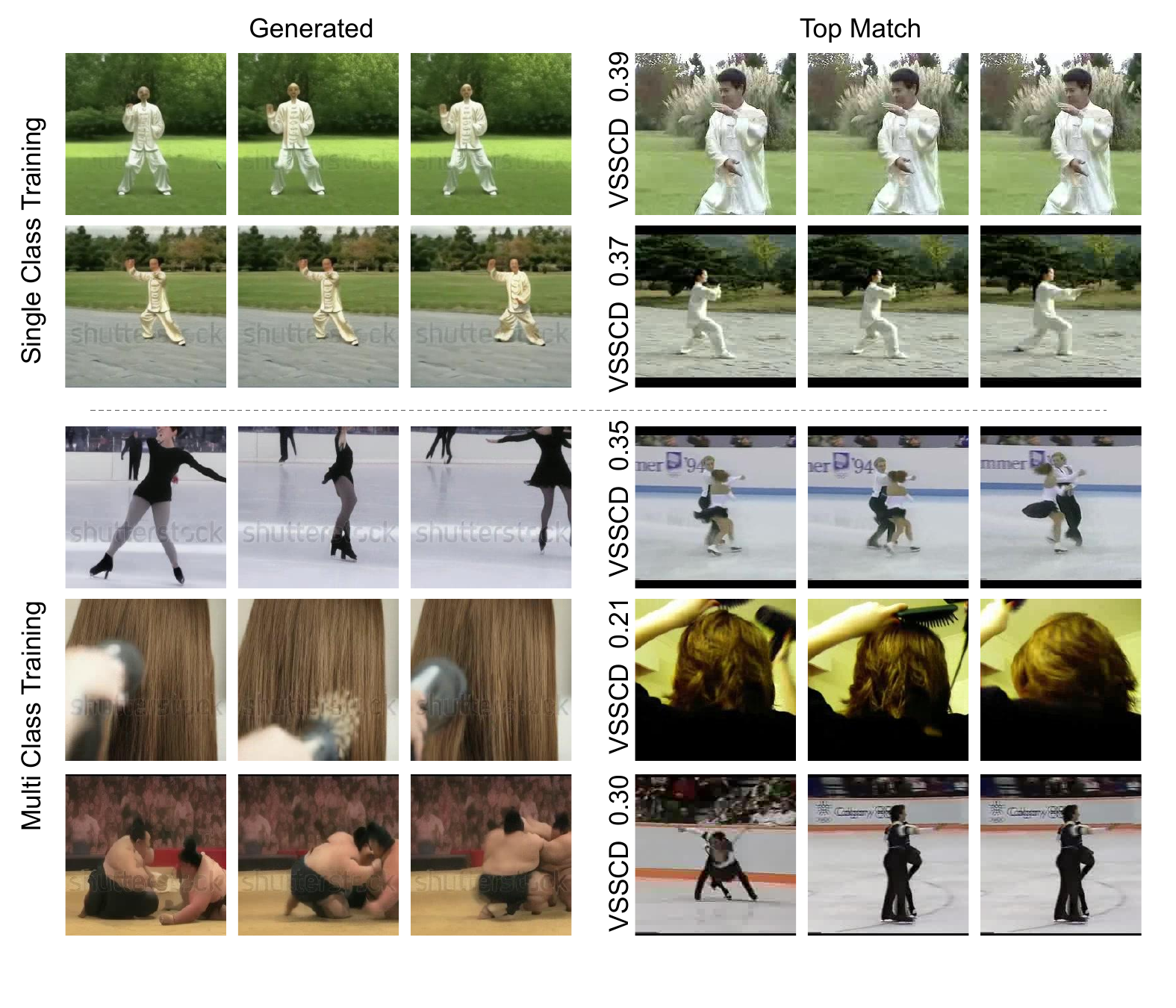}

\vskip-20pt\caption{We fine-tuned the temporal layers of a video diffusion model \cite{wang2023modelscope} using two different scenarios. In the first scenario, we trained the model with approximately 100 videos belonging to a single class. In the second scenario, we conducted multi-class training, utilizing around 1,000 different videos across 101 distinct classes. Notably, we observed that in both cases, the similarity between the generated videos and the training data was remarkably low, indicating that the model successfully produced unique video content.}
\label{Fig:tiv}
\vskip-6mm
\end{figure}

\subsection{Utilizing Text-to-Image Backbones}
\label{t2i}
In recent methods, several architectures have incorporated a Text-to-Image (T2I) foundation and augmented it with additional temporal layers, enabling them to generate videos \cite{luo2023videofusion,blattmann2023align,wang2023modelscope}. In this approach, the image generation model serves as a foundation for spatial context creation, while added temporal layers focus on learning motion dynamics. Models developed using this method don't rely exclusively on limited video datasets to learn both content and motion, potentially reducing their vulnerability to sample replication. However, it is worth noting that these backbone models, often variants of Stable diffusion \cite{Latentdiff} for text-to-image tasks, are not immune to replicating samples from their own image training data \cite{somepalli2023diffusion,somepalli2023understanding}. While acknowledging this limitation, our paper does not delve into that specific aspect.  Instead, we concentrate on instances of sample replication within the generated videos themselves, where either content or motion is directly mirrored from the video training data in text-to-video models. We utilized Stable Diffusion \cite{Latentdiff}, and expanded it with additional temporal layers \cite{wang2023modelscope}. We then trained the model with UCF-101 \cite{soomro2012ucf101}, which resulted in 47\% average top VSSCD score, proving to generate less replicated results.

\vspace{-1mm}
\subsection{Fine-tune only Temporal Layers.} 
\begin{figure}[h!]
\vspace{-10pt}
\centering
\includegraphics[width=0.7\linewidth,scale=1]
{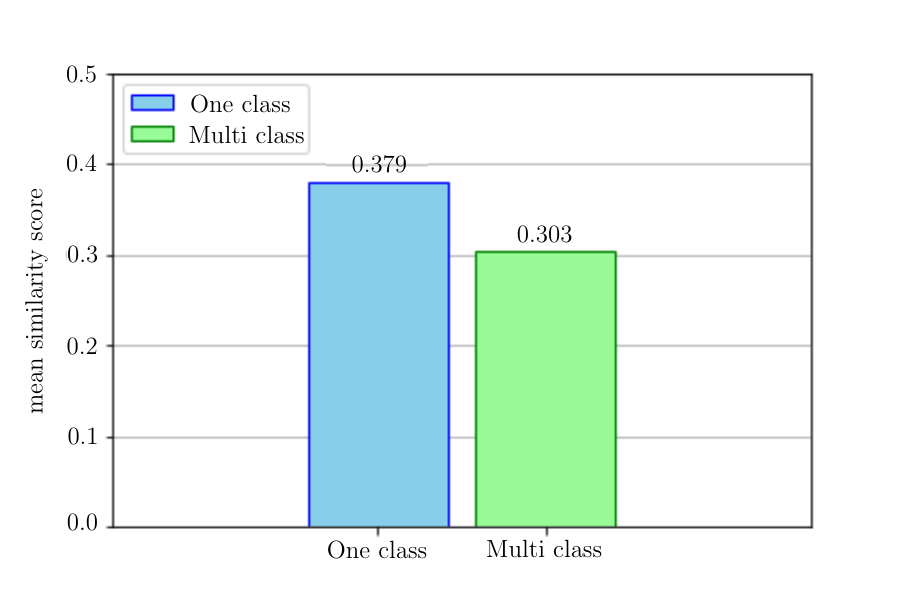}
\vskip-14pt
\caption{ Average highest VSSCD scores for video generation models trained on a single class versus all classes in the UCF-101 \cite{soomro2012ucf101} dataset. Both values fall significantly under 0.50, indicating that replication is not evident in the generated contents.}
\label{Fig:bar}
\vspace{-2mm}
\end{figure}

\vspace{-2mm}
As observed earlier, video diffusion models replicate existing content when trained on a small dataset. However, we found that this tendency is significantly reduced when only the temporal layers of the models are fine-tuned with video data. In our experiment, we specifically utilized Modelscope's video generation model \cite{wang2023modelscope} and fine-tuned its temporal layer using both the full UCF-101 dataset \cite{soomro2012ucf101} and its various individual classes. Notably, even with training on a relatively small sample size of around 100 examples, the model successfully steered clear of reproducing any content from the video training dataset, which is evident in Figure \ref{Fig:tiv}.  A comparison of average top VSSCD scores can be seen in Figure \ref{Fig:bar}. This success is largely due to the influence of the image generation backbone within the model, which directs the generation of new content. Based on these findings, we advocate for a similar fine-tuning approach, particularly when working with smaller video datasets.

\vspace{3mm}
\noindent\fbox{%
    \parbox{0.47\textwidth}{% Adjust the width as needed
        \textbf{\textcolor{blue}{Observation.}} Leveraging a Text-to-Image (T2I) backbone in video diffusion models demonstrates an enhanced ability to produce unique video content. Fine-tuning only the temporal layers of a pre-trained video diffusion model on a small dataset can effectively address the replication issue in low-resource settings.

    }%
}

\vspace{-3mm}
\section{Conclusion \& Future Work}
\vspace{-3mm}
In our study, we have conducted an in-depth examination of content and motion replication within video generation models. To the best of our knowledge, this is the first comprehensive analysis of replication in the context of video diffusion models. Our focus thus far has centered on the replication of both motion and content aspects. Our future research endeavors will pivot toward exploring the replication of motion across varying content. This involves the application of motion patterns, derived from training data, onto new content scenarios. Such an approach bears significance, particularly in contexts where motion patterns can be as distinctive as biometrics, posing potential risks.  Additionally, we aim to delve more deeper into models trained on extensive datasets. This is motivated by the precedent set by similar large-scale image models, which have demonstrated a tendency for replication. This future line of research promises to enrich our understanding of the nuances and capabilities inherent in video generation technology.

{\small
\bibliographystyle{ieee_fullname}
\bibliography{main}
}

\end{document}